\documentclass[conference]{IEEEtran}
\IEEEoverridecommandlockouts

\usepackage{cite}
\usepackage{amsmath,amssymb,amsfonts}
\usepackage{algorithmic}
\usepackage{graphicx}
\usepackage{textcomp}
\usepackage{xcolor}
\def\BibTeX{{\rm B\kern-.05em{\sc i\kern-.025em b}\kern-.08em
    T\kern-.1667em\lower.7ex\hbox{E}\kern-.125emX}}

\usepackage{caption}
\usepackage{listings}
\usepackage{xcolor}
\usepackage{paralist}
\usepackage{makecell,multirow}
\usepackage{rotating}
\usepackage{url}

\usepackage{times}
\usepackage{nohyperref}

\begin{document}

\title{\Large \textbf{ Large Scale Legal Text Classification Using Transformer Models}}

\author{\IEEEauthorblockN{Zein Shaheen}
\IEEEauthorblockA{
\textit{ITMO University}\\
St. Petersburg, Russia \\
shaheen@itmo.ru}
\and
\IEEEauthorblockN{Gerhard Wohlgenannt}
\IEEEauthorblockA{
\textit{ITMO University}\\
St. Petersburg, Russia \\
gwohlg@corp.ifmo.ru}
\and
\IEEEauthorblockN{Erwin Filtz}
\IEEEauthorblockA{
\textit{Vienna University of Economics and Business (WU)}\\
Vienna, Austria \\
erwin.filtz@wu.ac.at}}

\maketitle

\begin{abstract}
    Large multi-label text classification is a challenging Natural Language Processing (NLP) problem that is concerned with text classification for datasets with thousands of labels. 
    We tackle this problem in the legal domain, where datasets, such as JRC-Acquis and EURLEX57K labeled with the EuroVoc vocabulary were created 
    within the legal information systems of the European Union. The EuroVoc taxonomy includes around 7000 concepts. 
    In this work, we study the performance of various recent transformer-based models in combination with strategies such as generative pretraining, 
    gradual unfreezing and discriminative learning rates in order to reach competitive classification performance,
    and present new state-of-the-art results of $0.661$ (F1) for JRC-Acquis and $0.754$ for EURLEX57K.
    Furthermore, we quantify the impact of individual steps, such as language model fine-tuning
    or gradual unfreezing in an ablation study, and provide reference dataset splits created with an iterative stratification algorithm.
\end{abstract}

\begin{keywords}multi-label text classification; legal document datasets; transformer models; EuroVoc.\end{keywords}

\section{\uppercase{Introduction}}
\label{sec:introduction}


    Text classification, i.e., the process of assigning one or multiple categories from a set of options to a document~\cite{sebastiani2002machine}, 
    is a prominent and well-researched task in Natural Language Processing (NLP) and text mining. Text classification variants include simple binary classification (for example, decide if a document is spam or not spam), multi-class classification (selection of one from a number of classes), and multi-label classification. In the latter, multiple labels can be assigned to a single document.
    In \emph{Large Multi-Label Text Classification (LMTC)}, the label space is typically comprised of thousands of labels, which obviously raises task complexity. The work presented here tackles an LMTC problem in the legal domain.
    
    LMTC tasks often occur when large taxonomies or formal ontologies are used as document labels,
    for example in the medical domain~\cite{mullenbach2018explainable}~\cite{rios2018few}, 
    or when using large open domain taxonomies for labelling, such as annotating Wikipedia with labels~\cite{partalas2015lshtc}.
    A common feature of many LMTC tasks is that some labels are used frequently, while others are used very rarely (few-shot learning) or are never used (zero-shot learning). This situation is also referred to by \emph{power-law} or \emph{long-tail} frequency distribution of labels, which also characterizes our datasets and which is a setting that is largely unexplored for text classification~\cite{rios2018few}.
    Another difficulty often faced in LMTC datasets~\cite{rios2018few} are long documents, where finding the relevant areas 
    to correctly classify documents is a needle in a haystack situation.
    
    In this work, we focus on LMTC in the legal domain, based on two datasets, the well-known JRC-Acquis dataset~\cite{mencia2010} and
    the new EURLEX57K dataset~\cite{chalkidis2019large}. Both datasets contain legal documents from Eur-Lex
    \cite{eurLexSite}, the legal database of the European Union (EU).
    The usage of language in the given documents is highly domain specific, and includes many legal text artifacts such as case numbers. 
    Modern neural NLP algorithms often tackle domain specific text by fine-tuning pretrained language models on the type of text at hand~\cite{ruder2019transfer}.
    Both datasets are labelled with terms from the the European Union's multilingual and multidisciplinary thesaurus \emph{EuroVoc}
    \cite{eurVocThesaurus}.

    The goal of this work is to advance the state-of-the-art in LMTC based on these two datasets which exhibit many of the characteristics 
    often found in LMTC datasets: power-law label distribution, highly domain specific language and a large and hierarchically organized set of labels. 
    We apply current NLP transformer models, namely BERT~\cite{devlin-etal-2019-bert}, RoBERTa~\cite{liu2019roberta}, 
    DistilBERT~\cite{sanh2019distilbert}, XLNet~\cite{yang2019xlnet} and M-BERT~\cite{devlin-etal-2019-bert},
    and combine them with a number of training strategies such as gradual unfreezing, slanted 
    triangular learning rates and language model fine-tuning. 
    In the process, we create new standard dataset splits for JRC-Acquis and EURLEX57 
    using an iterative stratification approach~\cite{sechidis2011stratification}. Providing a high-quality standardized dataset split is very important, as previous work was typically done on different random splits, which makes results hard to compare~\cite{chang2019x}.
    Further, we make use of the semantic relations inside the EuroVoc taxonomy to infer reduced label sets for the datasets. 
    Some of our main evaluation results are the Micro-F1 score of $0.661$ for JRC-Acquis and $0.754$ for EURLEX57K, which sets new states-of-the-art to the best of our knowledge.
    
    The main findings and contributions of this work are: (i) the experiments with BERT, RoBERTa, DistilBERT, XLNet, M-BERT (trained on three languages), and AWD-LSTM in combination with the training 
    tricks to evaluate and compare the performance of the models, (ii) providing new standardized datasets for further investigation, 
    (iii) ablation studies to measure the impact and benefits of various training strategies, 
    and (iv) leveraging the EuroVoc term hierarchy to generate variants of the datasets for which higher classification
    performance can be achieved.
    
    The remainder of the paper is organized as follows: After a discussion of related work in Section~\ref{sec:related},
    we introduce the EuroVoc vocabulary and the two datasets (Section~\ref{sec:datasets}), and then present the main methods (AWD-LSTM,
    BERT, RoBERTa, DistilBERT, XLNet) in Section~\ref{sec:methods}.
    Section~\ref{sec:evaluation} contains extensive evaluations of the methods on both datasets as well as ablation studies,
    and after a discussion of results (Section~\ref{sec:dissc}) we conclude the paper in Section~\ref{sec:concl}.

%
%
                                

\section{Related Work}
\label{sec:related}
    
    In connection with the \emph{JRC-Acquis} dataset, Steinberger et al.~\cite{steinberger2013jrc} present the ``JRC EuroVoc Indexer JEX'',
    by the Joint Research Centre (JRC) of the European Commission.
    The tool categorizes documents using the EuroVoc taxonomy by employing
    a profile-based ranking task; the authors report an F-score between 0.44 and 0.54 depending on the document language.
    Boella et al.~\cite{boella2015linking} manage to apply a support vector machine approach to the problem by transforming the multi-label classification problem into a single-label problem. 
    Liu et al.~\cite{liu2017deep} present a new family of Convolutional Neural Network (CNN) models tailored for multi-label text classification.
    They compare their method to a large number of existing approaches on various datasets; for the EurLex/JRC dataset however, another method
    (SLEEC), provided the best results. SLEEC (Sparse Local Embeddings for Extreme Classification)~\cite{bhatia2015sparse}, creates
    local distance preserving embeddings which are able to accurately predict infrequently occurring (tail) labels. 
    The results on precision for SLEEC applied in Liu et al.~\cite{liu2017deep} are P@1: 0.78, P@3: 0.64 and P@5: 0.52 -- however, they use a 
    previous version of the JRC-Acquis dataset with only 15.4K documents.
    
    Chalkidis et al.~\cite{chalkidis2019large} recently published their work on the new EURLEX57K dataset. 
    The dataset will be described in more detail (incl.~dataset statistics) in the next sections.
    Chalkidis et al.~also provide a strong baseline for LMTC on this dataset.
    Among the tested neural architectures operating on the full documents, they have best results with BIGRUs with label-wise attention. 
    As input representation they use either GloVe~\cite{pennington2014glove} embeddings trained on domain text, 
    or ELMO embeddings~\cite{peters-etal-2018-deep}.
    The authors investigated using only the first zones of the (long) documents for classification, and show that 
    the title and recitals part of each document leads to almost
    the same performance as considering the full document~\cite{chalkidis2019large}. 
    This helps to alleviate BERT's limitation of having a maximum of 512 tokens as input.
    Using only the first 512 tokens of each document as input, BERT~\cite{devlin-etal-2019-bert} archives the best performance overall.
    The work of Chalkidis et al.~is inspired by You et al.~\cite{you2018attentionxml} who experimented with RNN-based methods with self attention 
    on five LMTC datasets (RCV1, Amazon-13K, Wiki-30K, Wiki-500K, and EUR-Lex-4K). 
    Similar work has been done in the medical domain, Mullenbach et al.~\cite{mullenbach2018explainable} investigate
    label-wise attention in LMTC for medical code prediction (on the MIMIC-II and MIMIC-III datasets).
    
     In this work, we experiment with BERT, RoBERTa, DistilBERT, XLNet, 
    M-BERT and AWD-LSTM. We provide ablation studies to measure the impact of various training strategies and heuristics. 
    Moreover, we provide new standardized datasets for further investigation by the research community, 
    and leverage the EuroVoc term hierarchy to generate variants of the datasets.
    
\section{Datasets and EuroVoc Vocabulary}
\label{sec:datasets}

    In this section, we first introduce the multilingual EuroVoc thesaurus which is used to classify legal documents published by the institutions of the European Union. The EuroVoc thesaurus is also used as a classification schema for the documents contained in the two legal datasets we use for our experiments, the \emph{JRC-Acquis V3} and \emph{EURLEX57K} datasets which are described in this section.

    \subsection{EuroVoc} \label{sec:eurovoc}
    \captionsetup{font={footnotesize},justification=centering,labelsep=period}
    \begin{figure}[t!]
    \begin{lstlisting}
    @prefix rdf: <http://www.w3.org/1999/02/22-rdf-syntax-ns#type> .
    @prefix skos: <http://www.w3.org/2004/02/skos/core#> .
    @prefix dcterms: <http://purl.org/dc/terms/> .
    @prefix ev: <http://eurovoc.europa.eu/> .
    @prefix evs: <http://eurovoc.europa.eu/schema#> .
    <http://eurovoc.europa.eu/100142>
        rdf:type evs:Domain ;
        skos:prefLabel "04 POLITICS"@en .
    <http://eurovoc.europa.eu/100166>
        rdf:type evs:MicroThesaurus ;
        skos:prefLabel "0421 parliament"@en ;
        dcterms:subject ev:100142 ;
        skos:hasTopConcept ev:41 .
    <http://eurovoc.europa.eu/41>
        rdf:type evs:ThesaurusConcept ;
        skos:prefLabel "powers of parliament"@en ;
        skos:inScheme ev:100166 .
    <http://eurovoc.europa.eu/1599>
        rdf:type evs:ThesaurusConcept ;
        skos:prefLabel "legislative period"@en ;
        skos:inScheme ev:100166
        skos:broader ev:41 .
    \end{lstlisting}
    \caption{EuroVoc example}
    \label{lst:eurovoc_example}
    \end{figure}

The datasets we use for our experiments contain legal documents from the legal information system of the European Union (Eur-Lex) and are classified into a common classification schema, the EuroVoc\cite{eurVocThesaurus} thesaurus published and maintained by the Publications Office of the European Union since 1982. The EuroVoc thesaurus has been introduced to harmonize the classification of documents in the communications across EU institutions and to enable a multilingual search as the thesaurus provides all its terms in the official language of the EU member states. 
It is organized based on the \emph{Simple Knowledge Organization System (SKOS)}\cite{w3cSkos}
, which encodes data using the \emph{Resource Description Format (RDF)}
\cite{w3cRDF}
and is well-suited to represent hierarchical relations between terms in a thesaurus like EuroVoc. 
EuroVoc uses SKOS to hierarchically organize its concepts into 21 domains, 
for instance \emph{Law, Trade} or \emph{Politics}, to name a few. 
Each domain contains multiple microthesauri (127 in total),
which in turn have in total around 600 top terms. 
About 7K terms (also called \emph{descriptors, concepts} or \emph{labels}) are assigned to one or multiple microthesauri and
connected to top terms using the predicate \path{skos:broader}.

All concepts in EuroVoc have a \emph{preferred} (\path{skos:prefLabel}) label and \emph{non-preferred} (\path{skos:altLabel}) label for each language; the label language is indicated with language tags. 
Figure~\ref{lst:eurovoc_example} illustrates with an example serialized in Turtle (TTL)
\cite{w3cTurtule}
format how the terms are organized in the EuroVoc thesaurus. 
Our example is from the domain \emph{04 POLITICS} and we show only the English labels of the concepts. The domain \emph{04 POLITICS} has the EuroVoc ID \path{ev:100142} and is of \path{rdf:type} \path{evs:Domain}. 
Each domain has microthesauri as the next lower level in the hierarchy. 
In this example, we can see that a \path{evs:Microthesaurus} named \emph{0421 parliament} is assigned to the \emph{04 POLITICS} domain using (\path{dcterms:subject} \path{ev:100142}) 
and is also connected to the next lower level of top terms.
The top term \emph{powers of parliament} (\path{ev:41}) is linked to the microthesaurus using \path{skos:inScheme}.
Finally, the lowest level in this example is the concept \emph{legislative period} (\path{ev:1599}) which is linked to its (\path{skos:broader}) top term \emph{powers of parliament} (\path{ev:41}), and is also directly linked to the microthesaurus \emph{0421 parliament} to which it belongs to using \path{skos:inScheme}.

The legal documents are annotated with multiple EuroVoc classes typically on the lowest level which results in a huge amount of available classes a document can be potentially classified in. In addition, this also comes with the disadvantage of the power-law distribution of labels such that some labels are assigned to many documents whereas others are only assigned to a few documents or to no documents at all.
The advantages of using a multilingual and multi-domain thesaurus for document classification are manifold.
Most importantly, it allows us to reduce the numbers of potential classes by going up the hierarchy, which does not make classification incorrect but only more general. Reducing the number of labels allows to compare the efficiency of the model for different label sets, which vary in size and sparsity.
In this line, we use a class reduction method to generate datasets with a reduced number of classes by replacing the original labels with the \emph{top terms}, \emph{microthesauri} or \emph{domains} they belong to.  For the top terms dataset, we leverage the \path{skos:broader} relations of the original descriptors,
for the microthesauri dataset we follow \path{skos:inScheme} links to the microthesauri, and 
the domains dataset is inferred via the \path{dcterms:subject} links of the microthesauri.
This process creates three additional datasets 
(\emph{top terms, microthesauri, domains})~\cite{filtz2019exploiting}. 
Furthermore, such a thesaurus would also allow to incorporate potentially more fine-grained national thesauri of member states which could be aligned with EuroVoc and therefore enable multilingual search in an extended thesarus.

%
%

    \subsection{Legal Text Datasets}
    
        In this work we focus on legal documents collected from the Eur-Lex
        \cite{eurLexSite} database serving as the official site for retrieving European Union law, such as \emph{Treaties}, \emph{International agreements} and \emph{Legislation}, and case law of the European Union (EU).
        Eur-Lex provides the documents in the official languages of the EU member states.
        As discussed in previous work \cite{filtz2019exploiting} the documents are well structured and written in domain specific
        language. Furthermore, legal documents are typically longer compared to texts often taken for text classification task such as the Reuters-21578 dataset containing news articles.

        In this paper, we use the English versions of the two legal datasets \emph{JRC-AcquisV3} \cite{jrcsaquis}
        and \emph{EURLEX57K}
        \cite{eurlex57k}. The \emph{JRC-Acquis V3} dataset has been compiled by the Joint Research Centre (JRC) of the European Union with the \emph{Acquis Communautaire} being the applicable EU law and contains documents in XML format. Each JRC document is divided into body, signature, annex and descriptors. The \emph{EURLEX57K} dataset has been prepared by academia~\cite{chalkidis2019large} and is provided in JSON format structured into several parts, namely the header including title and legal body, recitals (legal background references), the main body (organized in articles) and the attachments (appendices, annexes). Furthermore and in contrast to JRC-Acquis, the EURLEX57K dataset is already provided with a split into train and test sets.

    Table~\ref{tab_stats} shows a comparison of the dataset characteristics. \emph{EURLEX57K} contains almost three times as many documents as the \emph{JRC-Acquis V3} dataset, but the documents are comparable in their minimum number of tokens, median and mode of tokens per document. 
    The large difference in the maximum number of tokens per document impacts the standard deviation and the mean number of tokens. The reason for this difference is that JRC-Acquis also includes documents dealing with the budget of the European Union, comprised of many tables. As both datasets originate from the same source, but with different providers, we analyzed the number of  documents contained in both datasets and found an overlap of approx.~12\%.
    
        
        \captionsetup{font={footnotesize,sc},justification=centering,labelsep=period}%
        \begin{table}[t]
\small
            \centering
            \caption{Dataset statistics for JRC-Acquis and EURLEX57K. 
            }
            \label{tab_stats}
            \begin{tabular}{|c|c|c|}
            \hline
              & JRC-Acquis & EURLEX57K \\\hline
            \#Documents         & 20382   &  57000  \\\hline
            Max \#Tokens/Doc    & 469820  & 3934    \\\hline
            Min \#Tokens/Doc    & 21      & 119     \\\hline
            Mean \#Tokens/Doc   & 2243.43 & 758.46  \\\hline
            StdDev \#Tokens/Doc & 7075.94 & 542.86  \\\hline
            Median \#Tokens/Doc & 651.0   & 544     \\\hline
            Mode \#Tokens/Doc   & 275     & 275     \\\hline
            \end{tabular}
        \end{table}
        \captionsetup{font={footnotesize,rm},justification=centering,labelsep=period}%
        
                Table~\ref{tab_labels_stats} provides an overview of label statistics for both datasets. We created different versions based on the original descriptors (DE), top terms (TT), microthesauri (MT) and domains (DO) and present the numbers for all versions. 
        The maximum number of labels assigned to a single document is similar 
        for both datasets. The average number of labels per document in the original (DE) version is $5.46$ (JRC-Acquis) and $5.07$ (EURLEX57). Due to the polyhierarchy in the geography domain a label may be assigned to multiple \emph{Top Terms}, therefore 
        the number of \emph{Top Term} labels is higher than that of the original descriptors.
        
        
        \captionsetup{font={footnotesize,sc},justification=centering,labelsep=period}%
        \begin{table*}[t]
\small
            \centering
            \caption{Dataset statistics -- number of labels per document.}
            \label{tab_labels_stats}
            \begin{tabular}{|c|c|c|c|c|c|c|c|c|}
            \hline
              & \multicolumn{4}{c|}{JRC-Acquis} & \multicolumn{4}{c|}{EURLEX57K} \\\hline 
Label             & DE  & TT & MT & DO & DE & TT & MT & DO \\\hline
Max        &  24    & 30    & 14    & 10   &  26   & 30   & 15   &  9    \\\hline
Min        &  1     & 1     & 1     & 1    &  1    & 1    & 1    &  1    \\\hline
Mean       &  5.46  & 6.04  & 4.74  & 3.39 &  5.07 & 5.94 & 4.55 &  3.24 \\\hline
StdDev     &  1.73  & 3.14  & 1.92  & 1.17 &  1.7  & 3.06 & 1.82 &  1.04 \\\hline
Median     &  6     & 5     &  5    & 3    &  5    & 5    & 4    &  3    \\\hline
Mode       &  6     & 4     &  4    & 3    &  6    & 4    & 4    &  3    \\\hline

            \end{tabular}
        \end{table*}
        \captionsetup{font={footnotesize,rm},justification=centering,labelsep=period}%

\begin{figure}[htbp]
            \centering
            \includegraphics[scale=.4]{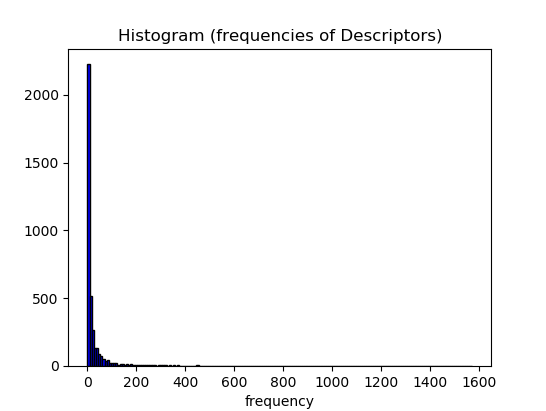}
            \caption{Power-law distribution of descriptors in the JRC-Acquis dataset.}
            \label{fig:histo}
        \end{figure}

        Figure~\ref{fig:histo} visualizes the power-law (long tail) label distribution,
        where a large portion of EuroVoc descriptors is used rarely (or never) as document annotations.
        In the JRC-Acquis dataset only 50\% of the labels available in EuroVoc are used to classify documents. Only 417 labels are used frequently (used on more than 50 documents) and 3,3147 labels have a frequency between 1--50 (few-short). 
        The numbers for the EURLEX57K dataset are similar \cite{chalkidis2019large}, 
        with 59.31\% of all EuroVoc labels being actually present in EURLEX57K. 
        From those labels, 746 are frequent, 3,362 have a frequency between 1--50,
        and 163 are only in the testing, but not in the training, dataset split (zero-shot). 
        The high number of infrequent labels obviously is a challenge when using supervised learning approaches.
        
        \captionsetup{font={footnotesize},justification=centering,labelsep=period}%
        \captionsetup{font={footnotesize,rm},justification=centering,labelsep=period}%

\section{Methods}
\label{sec:methods}

    In this section we describe the methods used in the LMTC experiments presented in the evaluation section,
    and the general training process.
    Furthermore, we discuss important related points such as language model pretraining and fine-tuning,
    and discriminative learning rates, and other important foundations for the evaluation section like dataset splitting
    and multilingual training.
    

    \subsection{General Training Strategy and Implementation} 
    
        In accordance with common NLP practice, as first introduced by Howard and Ruder for text classification~\cite{howard2018universal}, we train our
        models in two steps: first we fine-tune the language modeling part of the model to the target corpus (JRC-Acquis or EURLEX57K), and then we train the classifier on the training-split of the dataset. 
        
        The baseline model (AWD-LSTM) and the transformer models 
        are available with pretrained weights, 
        trained with language modelling objectives on large corpora such as Wikitext or Webtext -- a process that is computationally very expensive.
        Fine-tuning allows to transfer the language modeling capabilities to a new domain~\cite{howard2018universal}.
    
        Our implementation makes use of the FastAI library
        \cite{docsfastai}, which includes the basic infrastructure to apply training strategies like
        gradual unfreezing or slanted triangular learning rates (see below). 
        Moreover, for the transformer models, we integrate the Hugging Face 
        transformers package
        \cite{huggingfacetransformers} with FastAI.
       
        Our implementation including the evaluation results, is available on GitHub\cite{shaheenEtALRepository}. The repository also includes the reference
        datasets created with iterative splitting, 
        which can be used by other researchers as reference datasets -- in order to have a fair comparison of different approaches in the future.
       
        
        
    \subsection{Tricks for Performance Improvement (within FastAI)} 
        \label{sec:tricks}
 
        In their Universal Language Model Fine-tuning for Text Classification
        (ULMFiT) approach, Howard and Ruder~\cite{howard2018universal} propose a number of training strategies and tricks to improve model performance, which
        are available within the FastAI libary.
        Firstly, based on the idea that early layers in a deep neural network capture more general and basic features of language, which need little domain adaption,
        \emph{discriminative fine-tuning} applies different learning rates depending on the layer; earlier layers use smaller learning rates compared to later layers. 
        Secondly, \emph{slanted triangular learning rates} quickly increase the learning rate at the beginning of a training epoch up to the maximal learning rate
        in order to find a suitable region of the parameter space, and then slowly reduce the learning rate to refine the parameters. 
        And finally, in \emph{gradual unfreezing} the training process is divided into multiple cycles, 
        where each cycle consists of several training epochs.
        Training starts after freezing all layers except for the last few layers in cycle one, 
        during later cycles more layers are unfrozen gradually (from last to first layers). 
        The intuition is that, in fine-tuning a deep learning model (similar to discriminative fine-tuning), 
        that later layers are more task and domain specific and need more fine-tuning.
        In the evaluation section, we provide details about our unfreezing strategy (Table~\ref{tab-gradual unfreezing}).
        
    \subsection{Baseline Model} 
    
        We use \textbf{AWD-LSTM}~\cite{merity2017regularizing} as a baseline model. 
        Merity et al.~\cite{merity2017regularizing} investigate different strategies 
        for regularizing word-level LSTM language models, including the \emph{weight-dropped LSTM}
        with its recurrent regularization, and they introduce NT-ASGD as a new version of 
        average stochastic gradient descent in AWD-LSTM.
        
        In the ULMFiT approach~\cite{howard2018universal} of FastAI, AWD-LSTM is used as encoder, 
        with extra layers added on top for the classification task.
        
        For any of the models (AWD-LSTM and transformers) we apply the basic method discussed above:
        a) fine-tune the language model on all documents (ignoring the labels) of the dataset (JRC-Acquis or EURLEX57K),
        and then b) fine-tune the classifier using the training-split of the dataset.
      
    \subsection{Transformer Models} 
        In the experiments we study the performance of BERT, RoBERTa, DistilBERT and XLNet on the given text
        classification tasks. BERT is an early, and very popular, transformer model, RoBERTa is a modified version of BERT trained on a larger corpus, DistilBERT is a distilled version of BERT and thereby with lower computational cost, 
        and finally, XLNet can be fed with larger input token sequences. 
     
        \textbf{BERT:} BERT \cite{devlin-etal-2019-bert} is a bidirectional language model which aims to learn contextual relations between words 
        using the transformer architecture~\cite{vaswani2017attention}. We use an official release of the pre-trained models, 
        details about the specific hyperparameters are found in Section~\ref{eval:setup}.
        
        The input to BERT is either a single text (a sentence or document), or a text pair.
        The first token of each sequence is the special classification token [CLS], followed by WordPiece tokens of the first text \textit{A}, then a separator token [SEP], and (optionally) after that WordPiece tokens for the second text \textit{B}. 
        
        In addition to token embeddings, BERT uses positional embeddings to represent the position of tokens in the sequence. 
        For training, BERT applies Masked Language Modeling (MLM) and Next Sentence Prediction (NSP) objectives. 
        In MLM, BERT randomly masks 15\% of all WordPiece tokens in each sequence and learns to predict these masked tokens. 
        For NSP, BERT is fed in 50\% of cases with the actual next sentence \textit{B}, in the other cases 
        with a random sentence \textit{B} from the corpus.
       
        \textbf{RoBERTa:} RoBERTa, introduced by Liu et al.~\cite{liu2019roberta}, retrains BERT with an improved methodology, much more data, 
        larger batch size  and longer training times. In RoBERTa the training strategy of BERT is modified by removing the NSP objective.   
        Further, RoBERTa uses byte pair encoding (BPE) as a tokenization algorithm instead of WordPiece tokenization in BERT.
  
        \textbf{DistilBERT:} We use a distilled version of BERT released by Sanh et al.~\cite{sanh2019distilbert}. 
        DistilBERT provides a lighter and faster version of BERT, reducing the size of the model by 40\% while retaining 97\% of its capabilities on language understanding tasks~\cite{sanh2019distilbert}.
        The distillation process includes training a complete BERT model (the teacher) using the improved methodology proposed by Liu et al.~\cite{liu2019roberta}, 
        then DistilBERT (the student) is trained to reproduce the behaviour of the teacher by using cosine embedding loss.
        
        \textbf{XLNet:} The previously discussed transformer-based models are limited to a fixed context length 
        (such as 512 tokens), while legal documents are often long and exceed this context length limit. 
        XLNet~\cite{yang2019xlnet} includes segments recurrence, introduced in Transformer-XL~\cite{dai2019transformer}, 
        allowing it to digest longer documents. XLNet follows RoBERTa in removing the NSP objective, while introducing a novel permutation language model objective.
        In our work with XLNet, we fine-tune the classifier directly without LM fine-tuning 
        (as LM fine-tuning of XLNet was computationally not possible on the hardware available for our experiments).

      \subsection{Dataset Splitting}
        \label{sec:ds_splitting}

        \emph{Stratification} of classification data aims at splitting the data in a way that
        in all dataset splits (training, validation, test) the target classes appear in similar proportions.
        In multi-label text classification \emph{stratification} becomes harder, 
        because the target is a combination of multiple labels.        
        In \emph{random splitting}, it is possible that most instances of a specific class
        end up either in the training or test split (esp.~for low frequency classes), 
        and therefore the split can be unrepresentative with respect to the original data set. 
        Moreover, random splitting and different train/validation/test ratios create 
        the problem that results from different approaches are hard to compare~\cite{chang2019x}.
        
        Depending on the dataset, other criteria can be used for dataset splitting, for example
        Azarbonyad et al.~\cite{azarbonyad2019many} split JRC-Acquis documents according to document's year, where
        older documents could be used in training, and newer in testing. 
        
        For splitting both JRC-Acquis and EURLEX57K, we use the iterative stratification algorithm
        proposed by Sechidis et al.~\cite{sechidis2011stratification}, ie.~its implementation provided 
        by the scikit-multilearn library
        \cite{stratification}. 
        Applying this algorithm leads to a better document split with respect to the target labels, 
        and in turn, helps with generalization of the results and allows for a fair comparison of different approaches.
        The reference splits of the dataset are available online\cite{shaheenEtALRepository}.
        
        In the experiments in Section~\ref{sec:evaluation} we use these dataset splits, but in addition for EURLEX57K 
        also the dataset split of the dataset creators~\cite{chalkidis2019large}, in order to compare to their evaluation
        results.

    \subsection{Multilingual Training}
        JRC-Acquis is a collection of parallel texts in 22 languages -- we make use of this property to train multilingual BERT\cite{bertMultiLingualRep} 
        on an extended version of JRC-Acquis in 3 languages. 
        Multilingual BERT provides support for 104 languages and it is useful for zero-shot learning tasks 
        in which a model is trained using data from one language and then used to make inference on data in other languages.
        
        We extend the English JRC-Acquis dataset with parallel data in German and French.
        The additional data has the same dataset split as in the English version, ie.~if an English document 
        is in the training set then the German and French versions will be in the same split as well.
        

\section{Evaluation}
\label{sec:evaluation}
    This section first discusses evaluation setup (for example model hyperparameters) and then evaluation results for JRC-Acquis and EURLEX57K.


    \subsection{Evaluation Setup}
        \label{eval:setup}
        
        Evaluation setup includes important aspects such as dataset splits, preprocessing, the specific model architectures and variants,
        and major hyperparameters used in training.
       
        \paragraph{Dataset Splits:} 
        The official JRC-Acquis dataset does not include a standard train-validation-test split, and as discussed in Section~\ref{sec:ds_splitting}
        a random split exhibits unfavorable characteristics. We apply iterative splitting~\cite{sechidis2011stratification} to ensure
        that each split has the same label distribution as the original data. We split with an 80\%/10\%/10\% ratio for training/validation/test sets. 
        For the EURLEX57K the dataset creators already provide a split and a strong baseline evaluation. We run our models on the given split in order to compare
        results,
        and also create our own split with iterative splitting (dataset available in the mentioned GitHub repository\cite{shaheenEtALRepository}).

                \paragraph{Text Preprocessing:} All described models have their own preprocessing included (e.g. WordPiece tokenization
        in BERT), we do not apply extra preprocessing to the text.
        
        \paragraph{Neural Network Architectures:} 
        For \textbf{AWD-LSTM}, we use the standard setup of the pretrained model included in FastAI, 
        which has an input embedding layer with embedding size of 400, followed by three LSTM layers with hidden sizes of 1152
        and weight dropout probability of 0.1. 

        For the transformer models, we start from pretrained models, 
        the uncased BERT model
        \cite{huggingfacebert},
        the RoBERTa model
        \cite{huggingfaceroberta}, 
        DistilBERT
        \cite{huggingfacedistilbert}, 
        and the XLNET model
        \cite{huggingfacexlnet}.
        
        \captionsetup{font={footnotesize,sc},justification=centering,labelsep=period}%
        \begin{table}[htbp]
\small
        \centering
        \caption{Architecture hyperparameters of transformer models}
        \label{tab-transformers}
        \settowidth\rotheadsize{Context}
        \begin{tabular}{|c|c|c|c|c|c|}
        \hline
         \rothead{Model Name}    & \rothead{\# Layers} & \rothead{\# Heads} & \rothead{Context Length} & \rothead{Is Cased} & \rothead{batch-size}\\\hline
         BERT         & 12       & 12                 & 512         & False    & 4         \\\hline
         Roberta      & 12       & 12                 & 512         & False    & 4         \\\hline
         DistilBERT   & 6        & 12                 & 512         & False    & 4         \\\hline
         XLNet        & 12       & 12                 & 1024        & True     & 2         \\\hline
         
        \end{tabular}
        \end{table}
        \captionsetup{font={footnotesize,rm},justification=centering,labelsep=period}%
        
        In Table~\ref{tab-transformers}, we see that many architectural details are similar for the different model types.
        The transformer models all have 12 network layers, except DistilBERT with 6 layers, and 12 attention heads.
        XLNet allows for longer input contexts, but for performance reasons we limited the context to 1024 tokens, 
        and it was necessary to reduce the batch size to $2$ to fit the model into GPU memory, and also we could not unfreeze 
        the whole pretrained model (see below).

        To create the text classifiers, we take the representation of the text generated by the transformer model or AWD-LSTM, 
        and add two fully connected layers of size 1200 and 50, respectively, with a dropout probability of 0.2, 
        and an output layer. We apply batch normalization on the fully connected layers.

       
        \paragraph{Gradual Unfreezing:} Gradual unfreezing is one of the ULMFiT strategies discussed in Section~\ref{sec:tricks},
        where the neural network layers are grouped, and trained starting with the last group, 
        then incrementally unfrozen and trained further.
        
        \captionsetup{font={footnotesize,sc},justification=centering,labelsep=period}%
        \begin{table}[htbp]
\small
        \centering
        \caption{Gradual unfreezing details: Learning Rates (LR), number of epochs (Iters), and layer groups that are unfrozen.}
        \label{tab-gradual unfreezing}
        \settowidth\rotheadsize{DistilBERT}
        \begin{tabular}{|c|c|c|c|c|c|}
        \hline
        &&& \multicolumn{3}{|c|}{\# Unfrozen Layers} \\\cline{4-6}
         Cycle & Max LR    &  \# Iters & \rothead{BERT RoBERTa} & \rothead{DistilBERT} & \rothead{XLNet} \\\hline
         1       & 2e-4  & 12  &  4      &  2 & 4    \\\hline
         2       & 5e-5  & 12  &  8      &  4 & 6    \\\hline
         3       & 5e-5  & 12  & 12      &  6 & 8    \\\hline
         4       & 5e-5  & 36  & 12      &  6 & 8    \\\hline
         5       & 5e-5  & 36  & 12      &  6 & 8    \\\hline
        \end{tabular}
        \end{table}
        \captionsetup{font={footnotesize,rm},justification=centering,labelsep=period}%

        Except for DistilBERT, which has only 2 layers per layer group, all transformer models have 3 groups of 4 layers used in
        the unfreezing process.
        Table~\ref{tab-gradual unfreezing} gives an overview of the training setup for the transformer models.
        We trained the classifier for 5 cycles, starting in cycle $1$ with $4$ layers and a $LR=2e-4$, 
        and $12$ training epochs (Iters). 
        The setup of the other cycles is shown in the table. Overall, we used the same setup for all transformer models
        with a goal of better comparison between models. (Remark: hand-picking LRs and training epochs might lead to slightly better results.)
        
        \captionsetup{font={footnotesize,sc},justification=centering,labelsep=period}%
        \begin{table}[htbp]
\small
        \centering
        \caption{Gradual unfreezing settings for AWD-LSTM}
        \label{tab-gradual unfreezing2}
        \begin{tabular}{|c|c|c|c|}
        \hline
        Cycle   &  \# Max LR   & \# Unfrozen Layers & \# Iterations \\\hline
         1       & 2e-1  & 1    & 2    \\\hline
         2       & 1e-2  & 2    & 5    \\\hline
         3       & 1e-3  & 3    & 5    \\\hline
         4       & 5e-3  & all  & 20    \\\hline
         5       & 1e-4  & all  & 32    \\\hline
         6       & 1e-4  & all  & 32    \\\hline
        \end{tabular}
        \end{table}
        \captionsetup{font={footnotesize,rm},justification=centering,labelsep=period}%
        
        Table~\ref{tab-gradual unfreezing2} shows the main hyperparameters of AWD-LSTM training, we trained the model
        in $6$ cycles, with LRs, epochs per cycle, and unfrozen layers as shown in the table.

\paragraph{LM Fine-tuning:} For the transformer models we do LM fine-tuning for 5 iterations, 
        with a batch size of $4$ and LR of $5e-5$. Transformer fine-tuning is done with a 
        script\footnote{https://github.com/huggingface/transformers/blob/master /examples/language-modeling/run\_language\_modeling.py}
        provided by Hugging Face.
        For the AWD-LSTM model we first fine-tune the frozen LM for 2 epochs, and then in cycle two fine-tune the unfrozen model for another 5 epochs.


    \paragraph{Hardware specifications}
    We trained the models on a single GPU device (NVIDIA GeForce GTX 1080 with 11~GB of GDDR5X memory). For inference, we use an Intel i7-8700K CPU @ 3.70GHz and 16GB RAM.

        \subsection{Evaluation Metrics}    

        In the evaluations, in line with Chalkidis et al.~\cite{chalkidis2019large}, we apply the following evaluation metrics: \emph{micro-averaged F1, R-Precision@K (RP@K)}, and \emph{Normalized Discounted Cumulative Gain (nDCG@K)}.  
        \emph{Precision@K (P@K)} and \emph{Recall@K (R@K)} are popular measures in LTMC, too, but they unfairly penalize in situations where the number of gold labels is unequal to $K$, which is the typical situation in our datasets. This problem led to the introduction of more suitable metrics like RP@K and nDCG@K. In the following, we briefly discuss the metrics.
       
        The $F1$-score is a common metric in information retrieval systems, 
        and it is calculated as the harmonic mean between precision and recall. 
        If we have a label $L$, Precision, Recall, and $F1$-score with respect to $L$ are calculated as follows: 
        
        {\centering
        $Precision_L=\frac{TruePositives_L}{TruePositives_L+FalsePositives_L}$ \\
        
        $Recall_L=\frac{TruePositives_L}{TruePositives_L+FalseNegatives_L}$ \\
        
        $F1_L=2*\frac{Precision*Recall}{Precision+Recall}$ \\
        }
        
         \emph{Micro-F1} is an extension of the $F1$-score for multi-label classification tasks, and it treats the entire set of predictions as one vector and then calculates the $F1$. 
         We use grid search to pick the \emph{threshold} on the output probabilities of the models 
         that gives the best Micro-F1 score on the validation set. The threshold determines which labels we assign to the documents. 
        
         Propensity scores prioritize predicting a few relevant labels over the large number of irrelevant ones \cite{jain2016extreme}. R-Precision@K ($RP@K$) calculates precision for the top K ranked labels,  if the number of ground truth labels for a document is less than K, K is set to this number for this document. 

        {\centering
         $RP@K=\frac{1}{N}\sum_{n=1}^{N}\sum_{k=1}^{K}\frac{Rel(n,k)}{min(K,R_n)}$ \\
        }
         
        Where $N$ is the number of documents, $Rel(n,k)$ is set to 1 if the k-th retrieved label in the top-K labels of the n-th document is correct, otherwise it is set to 0 .  $R_n$ is the number of ground truth labels for the n-th document.
        
        Normalized Discounted Cumulative Gain \textit{nDCG@k} for the list of top $K$ ranked labels  measures ranking quality. It is based on the assumption that highly relevant documents are more useful than moderately relevant documents.

        {\centering
        $nDCG@K=\frac{1}{N}\sum_{n=1}^{N}Z_{k_n}\sum_{k=1}^{K}\frac{2^{Rel(n,k)}-1}{log_2(1+k)}$\\
        }
        $N$ is the number of documents, $Rel(n,k)$ is set to 1 if the k-th retrieved label in the top-K labels of the n-th document is correct, otherwise it is set to 0. $Z_{k_n}$ is a normalization factor to ensure $nDCG@K=1$ for a perfect ranking.
        
        \captionsetup{font={footnotesize,sc},justification=centering,labelsep=period}%
            \begin{table*}[htbp]
\small
    \centering
    \caption{Comparison between different transformer models, fine-tuned using the same number of iterations on JRC-Acquis.
    } 
    \label{tab:jrc:overview}
    \begin{tabular}{|c|c|c|c|c|c|c|}
    \hline
                 & BERT  & RoBERTa        & XLNet & DistilBERT   & AWD-LSTM & Multilingual BERT \\\hline
        Micro-F1 & 0.661 & 0.659          & 0.605 & 0.652          & 0.493    & \textbf{0.663} \\\hline
         RP@1    & 0.867 & 0.873          & 0.845 & \textbf{0.884} & 0.762    & 0.873\\\hline
         RP@3    & 0.784 & \textbf{0.788} & 0.736 & 0.78           & 0.619    & 0.783 \\\hline
         RP@5    & 0.715 & 0.716          & 0.661 & 0.711          & 0.548    & \textbf{0.717}\\\hline
         RP@10   & 0.775 & \textbf{0.778} & 0.733 & 0.775          & 0.627    & 0.777\\\hline
         nDCG@1  & 0.867 & 0.873          & 0.845 & \textbf{0.884} & 0.762    & 0.873\\\hline
         nDCG@3  & 0.803 & \textbf{0.807} & 0.762 & 0.805          & 0.651    & 0.804\\\hline
         nDCG@5  & 0.750 & \textbf{0.753} & 0.703 & 0.75           & 0.594    & 0.752\\\hline
         nDCG@10 & 0.778 & \textbf{0.781} & 0.746 & 0.779          & 0.630    & 0.780\\\hline
    \end{tabular}
    \end{table*}
    \captionsetup{font={footnotesize,rm},justification=centering,labelsep=period}%

\subsection{Evaluation Results}
    \label{eval:resuls}
    
    The evaluation results are organized into three subsections,
    results for the JRC-Acquis dataset, results for the EURLEX57K dataset,
    and finally results from ablation studies.
    
    
    \subsubsection{JRC-Acquis}      
 

    Table~\ref{tab:jrc:overview} presents an overview of the results on the JRC-Acquis dataset for the transformer
    models and the AWD-LSTM baseline, and initial results from the multilingual model.

    The observations here are as follows: Firstly, transformer-based models outperform 
    the LSTM baseline by a large margin. Further, within the transformer models RoBERTa and BERT
    yield best results, the scores are almost the same. 
    As expected, the distilled version of BERT is a bit lower in most metrics like Micro-F1, but the difference is small. 
    
    In this set of experiments, XLNet is behind DistilBERT, which we attribute to two main causes: 
    (i) for computational reasons (given the available GPU hardware), we could \emph{not} fine-tune the LM on XLNet, and in classifier training 
    we could \emph{not} unfreeze the full model. (ii) We used the same LR on all models; the choice of LR was influenced 
    by a recommendation on BERT learning rates in Devlin et al.~\cite{devlin-etal-2019-bert}, and may not be optimal for XLNet.
    Overall, we could not properly test XLNet due to its high computational requirements, 
    and did therefore not include it in the set of experiments on the EURLEX57K dataset.

    
    The initial set of experiments with multilingual BERT (M-BERT) provides very promising results, 
    on par with RoBERT and BERT. This is remarkable given the fact that we use the same amount of 
    global training steps -- which means, because our multilingual dataset is 3 times larger, 
    that on individual documents we train only a $1/3$ of the time. 
    We expect even better results with more training epochs. LM fine-tuning of the M-BERT model 
    was done on the text from all three languages (en, de, fr).
    
     
     Regarding comparisons to existing baseline results, firstly because of the problem of different dataset splits 
     (see Section~\ref{sec:ds_splitting}) results are hard to compare. However, Steinberger et al.~\cite{steinberger2013jrc}
     report an F1-score of $0.48$, Esuli et al.~\cite{esuli2019funnelling} report an F1 of $0.589$ and 
     Chang et al.~\cite{chang2019x} do not provide F1, but only P@5 (62.64) and R@5 (61.59).
     
    
    \captionsetup{font={footnotesize,sc},justification=centering,labelsep=period}%
    \begin{table}[htbp]
\small
    \centering
    \caption{BERT results for JRC-Acquis with \emph{class reduction} methods applied, which lead to 4 datasets:
    DE (descriptors), TT (top-terms), MT (microthesauri, DO (domains)}
    \label{tab:bert:jrc}
    \begin{tabular}{|c|c|c|c|c|}
    \hline
      & DE & TT & MT & DO \\\hline
    Micro-F1 & 0.661 & 0.745 & 0.778 & 0.839  \\\hline
    RP@1 & 0.867 & 0.922 & 0.943 & 0.967 \\\hline
    RP@3 & 0.784 & 0.838 & 0.871 & 0.905 \\\hline
    RP@5 & 0.715 & 0.804 & 0.844 & 0.928 \\\hline
    RP@10 & 0.775 & 0.857 & 0.908 & 0.974 \\\hline
    nDCG@1 & 0.867 & 0.922 & 0.943 & 0.967 \\\hline
    nDCG@3 & 0.803 & 0.858 & 0.888 & 0.919 \\\hline
    nDCG@5 & 0.750 & 0.829 & 0.864 & 0.929 \\\hline
    nDCG@10 & 0.778 & 0.852 & 0.896 & 0.952 \\\hline
    \end{tabular}
    \end{table}
    \captionsetup{font={footnotesize,rm},justification=centering,labelsep=period}%
    
    For Table~\ref{tab:bert:jrc}, we picked one transformer-based method, namely BERT, and analyzed its performance on
    the various JRC datasets resulting from \emph{class reduction} described in Section~\ref{sec:eurovoc}. 
    By using inference on the EuroVoc hierarchy, we created, additionally to the default descriptors dataset, 
    datasets for EuroVoc Top Terms (TT), Micro-Thesauri (MT), and EuroVoc Domains (DO). 
    With the reduced number of classes, classification performance is clearly rising, for example from a 
    Micro-F1 of 0.661 (descriptors) to 0.839 (EuroVoc domains).
    We argue that the results with the inferred labels show that our approach might be well-suitable for real-world
    applications in scenarios like automatic legal document classification or keyword/label suggestion --
    for example the RP@5 for domains (DO) is at 0.928, so the classification performance (depending on the use case requirements) 
    may be sufficient.
   
   \captionsetup{font={footnotesize},justification=centering,labelsep=period}%
        \begin{figure}[htbp]
            \centering
            \includegraphics[scale=.35]{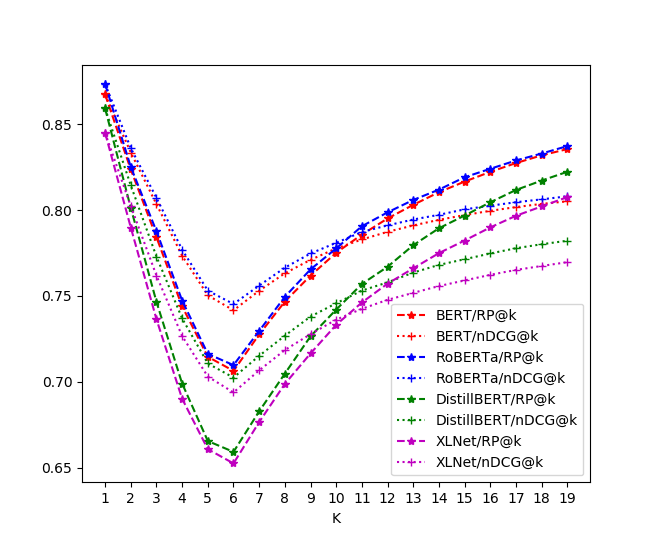}
            \caption{A visualization of RP@K and nDCG@K for all transformer models for JRC-Acquis.   }
            \label{fig:JRC_propensity}
        \end{figure}
    \captionsetup{font={footnotesize,rm},justification=centering,labelsep=period}%

        Figure~\ref{fig:JRC_propensity} contains a visual representation of RP@K and nDCG@K for the transformer models applied to 
        the JRC-Acquis dataset. We can see how similar the performance of BERT and RoBERTa is for different values of $K$, 
        and RoBERTa scores are consistently marginally better.
            

    \subsubsection{EURLEX57K}

         \captionsetup{font={footnotesize,sc},justification=centering,labelsep=period}%
        \begin{table*}[htbp]
\small
        \centering
        \caption{Results for our transformer-based models on EURLEX57K, and strong baselines from Chalkidis et al.}
        \label{tab:eurolex57k-overview}
        \begin{tabular}{|c|c|c|c||c|c|c|}
        \hline
          & \multicolumn{3}{c||}{Ours}& \multicolumn{3}{c|}{Chalkidis et al.~\cite{chalkidis2019large}}\\\hline
                    & BERT  & RoBERTa        & DistilBERT     & BERT-BASE       & BIGRU-LWAN-ELMO & BIGRU-LWAN-L2V \\\hline
        Micro-F1    & 0.751 & \textbf{0.758} & 0.754          & 0.732           & 0.719           & 0.709 \\\hline
        RP@1        & 0.912 & 0.919          & \textbf{0.925} &  0.922          & 0.921           & 0.915 \\\hline
        RP@3        & 0.843 & \textbf{0.85}  & 0.848          &  -              & -               & -  \\\hline
        RP@5        & 0.805 & \textbf{0.812} & 0.807          &  0.796          & 0.781           & 0.770 \\\hline
        RP@10       & 0.852 & 0.860          & \textbf{0.862} &  0.856          & 0.845           & 0.836 \\\hline
        nDCG@1      & 0.912 & 0.919          & \textbf{0.925} &  0.922          & 0.921           & 0.915 \\\hline
        nDCG@3      & 0.859 & \textbf{0.866} & \textbf{0.866} &  -              & -               & - \\\hline
        nDCG@5      & 0.828 & \textbf{0.835} & 0.833          &  0.823          & 0.811           & 0.801 \\\hline
        nDCG@10     & 0.849 & 0.857          & \textbf{0.858} &  0.851          & 0.841           & 0.832 \\\hline
        \end{tabular}
        \end{table*}
        \captionsetup{font={footnotesize,rm},justification=centering,labelsep=period}%
        
        %

        In this subsection we report the evaluation results on the new EURLEX57K dataset by Chalkidis et al.~\cite{chalkidis2019large}. 
        In order to compare to the results of the dataset creators, we ran the experiments on the dataset and dataset split 
        (45K training, 6K validation, 6K testing) provided by Chalkidis et al.~\cite{chalkidis2019large}. 
        Below, we also show evaluation results on our dataset split (created with the iterative stratification approach).
        Table~\ref{tab:eurolex57k-overview} gives an overview of results for our transformer models, and compares
        them to the strong baselines in existing work. 
        Chalkidis et al.~\cite{chalkidis2019large} evaluate various architectures,
        the results of the three best models presented here: BERT-BASE, BIGRU-LWAN-ELMO and BIGRU-LWAN-L2V.
        BERT-BASE is a BERT model with an extra classification layer on top, 
        BIGRU-LWAN combines a BIGRU encoder with Label-Wise Attention Networks (LWAN), 
        and uses either Elmo (ELMO) or word2vec (L2V) embeddings as inputs.
        Table~\ref{tab:eurolex57k-overview} shows that our models outperform the previous baseline, 
        the best results are delivered by RoBERTa and DistilBERT. The good performance of DistilBERT 
        in these experiments is surprising (We need further future experiments to explain the results sufficiently. One intuition might be that the random weight initialization of the added layers was very suitable.). 
        
        Overall, the results are much better than for the smaller JRC dataset, with the best Micro-F1 for JRC 
        being 0.661 (BERT), while for EURLEX57K we reach 0.758 (RoBERTa).

        \captionsetup{font={footnotesize,sc},justification=centering,labelsep=period}%
        \begin{table}[htbp]
\small                    
        \centering
        \caption{BERT results on EURLEX57K with \emph{class reduction} methods applied, plus
        the baseline results of BERT-BASE (DE) from Chalkidis et al.~\cite{chalkidis2019large}.}
        \label{tab:eurolex57k:bert}
        \settowidth\rotheadsize{baseline}
        \begin{tabular}{|c|c|c|c|c||c|}
        \hline
                    & DE    & TT     & MT     & DO    & \rothead{DE baseline} \\\hline
        Micro-F1    & 0.751 & 0.825  & 0.84   & 0.883 & 0.732  \\\hline
        RP@1        & 0.912 & 0.948  & 0.959  & 0.978 & 0.922 \\\hline
        RP@3        & 0.843 & 0.896  & 0.915  & 0.939 & - \\\hline
        RP@5        & 0.805 & 0.876  & 0.902  & 0.956 & 0.796 \\\hline
        RP@10       & 0.852 & 0.909  & 0.943  & 0.986 & 0.856 \\\hline
        nDCG@1      & 0.912 & 0.948  & 0.959  & 0.978 & 0.922\\\hline
        nDCG@3      & 0.859 & 0.907  & 0.924  & 0.947 & - \\\hline
        nDCG@5      & 0.828 & 0.891  & 0.912  & 0.955 & 0.823 \\\hline
        nDCG@10     & 0.849 & 0.904  & 0.931  & 0.97  & 0.851 \\\hline
        \end{tabular}
        \end{table}
        \captionsetup{font={footnotesize,rm},justification=centering,labelsep=period}%

        Table~\ref{tab:eurolex57k:bert} presents the results for BERT on the additional datasets with 
        Top Terms (TT), Micro-Thesauri (MT) and Domains (DO) labels inferred from the EuroVoc taxonomy
        (similar to Table~\ref{tab:bert:jrc}, which presents the scores of JRC-Acquis).
        As expected from the general results on the EURLEX57 dataset, the values on the derived datasets
        are better than for JRC-Acquis, for example RP@5 is now at $0.956$ for the domains (DO). 
        
        \captionsetup{font={footnotesize,sc},justification=centering,labelsep=period}%
         \begin{figure}[htbp]
            \centering
            \includegraphics[scale=.30]{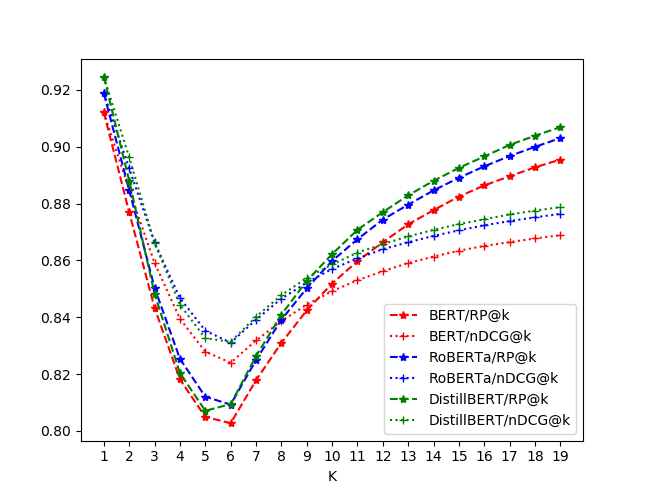}
            \caption{RP@K and nDCG@K for the transformer models trained on EURLEX57K. }
            \label{fig:EURLEX_propensity}
        \end{figure}
        \captionsetup{font={footnotesize,rm},justification=centering,labelsep=period}%

        Similar to Figure~\ref{fig:JRC_propensity},
        Figure~\ref{fig:EURLEX_propensity} shows RP@K and nDCG@K for BERT, RoBERTa and DistilBERT depending on the value of $K$.
        RoBERTa and DistilBERT are almost identical in their performance, BERT lags behind a little in this set of experiments. 

        Finally, in Table~\ref{tab:eurolex57kIterative}, we trained a BERT model on our iterative split of the EURLEX57K dataset in order to provide a strong baseline 
        for future work on a standardized and arguably improved version of the EURLEX57K dataset.

        \captionsetup{font={footnotesize,sc},justification=centering,labelsep=period}%
        \begin{table}[htbp]
\small
        \centering
        \caption{
        BERT results on EURLEX57K with the new iterative stratification dataset split.}
        \label{tab:eurolex57kIterative}
        \begin{tabular}{|c|c|c|c|c|}
        \hline
        Micro-F1 & RP@1 & RP@5 & nDCG@1 & nDCG@5 \\\hline 
        0.760 & 0.914 & 0.809 & 0.914 & 0.833 \\\hline
        
        \end{tabular}
        \end{table}
        \captionsetup{font={footnotesize,rm},justification=centering,labelsep=period}%
        
        \subsubsection{Ablation Studies}
        In this section, we want to study the contributions of various training process components 
        -- by excluding some of those components individually (or reducing the number of training epochs).
        We focus on three important aspects: (i) the use of Language Model (LM) fine-tuning, 
        (ii) gradual unfreezing, (iii) and a reduction of the number of training cycles.

        \captionsetup{font={footnotesize,sc},justification=centering,labelsep=period}%
        \begin{table}[htbp]
    \small
        \centering
        \caption{Classification metrics for the JRC-Acquis dataset, when \emph{not} using LM fine-tuning -- 
                in parentheses the results \emph{with} fine-tuning (for comparison).}
        \label{tab_lm_ft_abl}
        \begin{tabular}{|c|c|c|c|}
        \hline
                 & BERT          & RoBERTa       & DistilBERT    \\\hline
        Micro-F1 & 0.64 (0.66) & 0.65 (0.66) & 0.61 (0.62)  \\\hline
        RP@1     & 0.86 (0.87) & 0.87 (0.87) & 0.86 (0.87) \\\hline
        RP@3     & 0.77  (0.78) & 0.77 (0.79) & 0.75 (0.76) \\\hline
        RP@5     & 0.70 (0.72) & 0.70 (0.72) & 0.67 (0.68)  \\\hline
        RP@10    & 0.76 (0.78) & 0.77 (0.78) & 0.74 (0.75) \\\hline
        nDCG@1   & 0.86 (0.87) & 0.87 (0.87) & 0.86 (0.87) \\\hline
        nDCG@3   & 0.79 (0.80) & 0.79  (0.81) & 0.77 (0.78) \\\hline
        nDCG@5   & 0.74 (0.75) & 0.74 (0.75) & 0.71 (0.72) \\\hline
        nDCG@10  & 0.77 (0.72) & 0.77 (0.78) & 0.75 (0.76) \\\hline
        \end{tabular}
        \end{table}
        \captionsetup{font={footnotesize,rm},justification=centering,labelsep=period}%
 
        In Table~\ref{tab_lm_ft_abl}, we compare the evaluation metrics when removing the LM fine-tuning
        (on the legal target corpus) step before classification model training to the original version including
        LM fine-tuning (in parenthesis). For all examined models, we can see a small but consistent improvement of results
        when using LM fine-tuning. The relative improvement in the metrics is in the range of 1\%--3\%.
        In conclusion, LM fine-tuning to the legal text corpus is a crucial step for reaching a high classification performance.

 
 
       \captionsetup{font={footnotesize,sc},justification=centering,labelsep=period}%
        \begin{table}[htbp]
\small
        \centering
        \caption{Ablation study: BERT and DistilBERT performance on JRC-Acquis regarding the number of training epochs (Iter.)
        and the use of Gradual Unfreezing (GU).}
        \label{tab_gu_abl}
        \settowidth\rotheadsize{BERT}
        \begin{tabular}{|c|c|c|c|c|c|}
        \hline
        & \# Iter. & Use GU & Prec. & Rec. & Mic.-F1 \\\hline 
        & 36 & True & 0.678 &  0.601 & 0.637  \\\cline{2-6}
        & 108 & False & 0.674 & 0.575 & 0.621 \\\cline{2-6}
        \multirow[t]{-2}{*}{\rothead{BERT}}
        & 108 & True & 0.695 & 0.630 & \textbf{0.661} \\\cline{2-6}
        \hline

        & 36  & True & 0.696   & 0.601 & 0.645 \\\cline{2-6}
        & 108 & False &  0.663  & 0.583 & 0.620\\\cline{2-6}
        \multirow[t]{-2}{*} {\rothead{DistilBERT}}
        & 108 & True &  0.701  & 0.611 & 0\textbf{.653}
        \\\hline
        \end{tabular}
        \end{table}
        \captionsetup{font={footnotesize,rm},justification=centering,labelsep=period}%

        
        In Table~\ref{tab_gu_abl}, we examine the effect of two factors, the training epochs (Iter.) hyperparameter,
        and of the use of the gradual unfreezing technique. Regarding number of epochs, both models benefit from longer training,
        for BERT the difference is large (about 4\% relative improvement in F1-score), while for the simpler DistilBERT model less 
        training appears to be required, after 36 epochs it even provides better accuracy than BERT at this point, 
        and finally only gains a 1.2\% improvement from more training epochs. 
        Secondly, we study the effect of Gradual Unfreezing (GU), which for BERT has a large impact, with a relative
        improvement in F1 of about 6\%. 
        In summary, longer training times benefit esp.~more complex models like BERT, 
        and gradual unfreezing is a very helpful strategy for optimizing performance.

\section{Discussion}
\label{sec:dissc}

Much of the detailed discussion is already included in the \emph{Evaluation Results} section (Section~\ref{eval:resuls}),
so here we will summarize and extend on some of the key findings.

In comparing model performance, starting with LSTM versus transformer architectures,  
the results show that the attention mechanism used in transformers is superior to LSTMs in finding aspects relevant 
for the classification task in long documents. Within the transformer models, firstly we did not notice much difference between
BERT and RoBERTa, which is not unexpected, as they are technically very similar. 
Overall, results were a bit better for RoBERTa. 
DistilBERT delivered surprisingly good results for the EURLEX57K dataset, and has the benefits of lower computational cost. Both for the JRC-Aquis and the EURLEX57K datasets, 
the results indicate that DistilBERT is better in retrieving the most probable label compared 
with RoBERTa and BERT.
XLNet on the other hand, requires a lot of computational resources, and we were not able to properly train the model
for that reason. Finally, the first set of experiments on multilingual training with M-BERT gave promising results, 
hence it will be further studied in future work.

The ablation studies showed the positive effects of the training (fine-tuning) strategies that we applied, 
both LM-finetuning on the target domain, as well as gradual unfreezing of the network layers (in groups) 
proved to be crucial in reaching state-of-the-art classification performance.

To compare the computational costs, we calculated inference times for each model on an Intel i7-8700K CPU @ 3.70GHz. DistilBERT provides the lowest run time at 12 ms/example. 
RoBERTa and BERT (which have an identical architecture) have very similar run times with 17.1 ms, and 17.3 ms/example, respectively. 
XLNet, the heaviest model, requires 77 ms/example.

For a fair comparison, we trained all transformer models with the same set of hyperparameters 
(such as learning rate and number of training epochs). With customized and hand-picked parameters for each training 
cycle we expect further improvements of scores, which will be studied in future work together 
with model ensemble approaches and text data augmentation.

\section{Conclusions}
\label{sec:concl}

Natural Language Processing ( In) this work we evaluate current transformer models for natural language processing in combination with training strategies like
language model (LM) fine-tuning, slanted triangular learning rates and gradual unfreezing in the field of LMTC (large multi-label text classification)
on legal text datasets with long-tail label distributions. The datasets contain around 20K documents (JRC-Acquis) and 57K documents (EUROLEX57K) 
and are labeled with EuroVoc descriptors from the 7K terms in the EuroVoc taxonomy. 
The use of an iterative stratification algorithm for dataset splitting (into training/validation/testing) allows to create standardized splits 
on the two datasets to enable comparison and reproducibility in future experiments. In the experiments, we provide new state-of-the-art results
on both datasets, with a micro-F1 of $0.661$ for JRC-Acquis and $0.754$ for EUROLEX57K, and even higher scores for new datasets with reduced 
label sets inferred from the EuroVoc hierarchy (\emph{top terms, microthesauri, and domains}).

The main contributions are: 
(i) new state-of-the-art LMTC classification results on both datasets for a problem type that is still largely unexplored~\cite{rios2018few},
(ii) a comparison and interpretation of the performance of the applied models: AWD-LSTM, BERT, RoBERTa, DistilBERT and XLNet,
(iii) the creation and provision (on GitHub) of new standardized versions of the two legal text datasets created 
with an iterative stratification algorithm,
(iv) deriving new datasets with reduced label sets via the semantic structure within EuroVoc, and
(v) ablation studies that quantify the contributions of individual training strategies and hyperparameters such as 
gradual unfreezing, number of training epochs and LM fine-tuning in this complex LMTC setting.

There are multiple angles for \emph{future work}, including potentially deriving higher performance by using hand-picked learning rates
and other hyperparameters for each model individually, and further experiments on using models such as multilingual BERT to profit from
the availability of parallel corpora. Moreover, experiments with new architectures such as Graph Neural Networks~\cite{pal2020graphNN} 
and various data augmentation techniques are candidates to improve classification performance.

\section*{\uppercase{Acknowledgements}}
\noindent This work was supported by the Government of the Russian Federation (Grant 074-U01) through the ITMO Fellowship and Professorship Program.

\bibliographystyle{IEEEtran}
\bibliography{bibtemplate}

\end{document}